\documentclass[10pt,twocolumn,letterpaper]{article}

\usepackage[pagenumbers]{cvpr} 

\usepackage[dvipsnames]{xcolor}

\definecolor{cvprblue}{rgb}{0.21,0.49,0.74}
\usepackage[pagebackref,breaklinks,colorlinks,citecolor=cvprblue]{hyperref}

\usepackage{graphicx}
\usepackage{amsmath}
\usepackage{amssymb}
\usepackage{booktabs}

\usepackage{algorithm}
\usepackage{algpseudocode}
\usepackage{colortbl}
\usepackage{xcolor}
\usepackage{color}
\usepackage{amsmath}
\usepackage{lipsum}

\usepackage{wrapfig}
\usepackage[accsupp]{axessibility}

\usepackage{censor}
\usepackage{tikz}
\usepackage{pgfplots}
\usepackage{svg}
\usetikzlibrary{pgfplots.groupplots}
\usepgfplotslibrary{fillbetween}
\usepackage{adjustbox}
\usepackage{indentfirst}
\usepackage{tabularx}
\usepackage{bm}
\usepackage{svg}

\usepackage{comment}
\usepackage{censor}
\usepackage{tikz}
\usepackage{pgfplots}
\usepackage{svg}
\usetikzlibrary{pgfplots.groupplots}
\usepgfplotslibrary{fillbetween}
\usepackage{adjustbox}
\usepackage{indentfirst}
\usepackage{tabularx}

\usepgfplotslibrary{groupplots}

\usepackage{float}

\newcommand\blfootnote[1]{%
  \begingroup
  \renewcommand\thefootnote{}\footnote{#1}%
  \addtocounter{footnote}{-1}%
  \endgroup
}

\def\paperID{2} 
\def\confName{CVPRW}
\def\confYear{2024}

\title{On the Application of Egocentric Computer Vision to Industrial Scenarios}

\author{Vivek Chavan*\textsuperscript{1} \and
Oliver Heimann\textsuperscript{1} \and 
Jörg Krüger\textsuperscript{1,2} \vspace{0.2cm}\and 
\textsuperscript{1} Fraunhofer IPK \hspace{1cm} \textsuperscript{2} Technical University of Berlin, Germany\\
}

\begin{document}
\maketitle
\begin{abstract}

Egocentric vision aims to capture and analyse the world from the first-person perspective. We explore the possibilities for egocentric wearable devices to improve and enhance industrial use cases w.r.t. data collection, annotation, labelling and downstream applications. This would contribute to easier data collection and allow users to provide additional context. We envision that this approach could serve as a supplement to the traditional industrial Machine Vision workflow. Code, Dataset and related resources will be available at: \texttt{\href{https://github.com/Vivek9Chavan/EgoVis24}{https://github.com/Vivek9Chavan/EgoVis24}}

\end{abstract}    
\section{Introduction}
\label{sec:intro}

\blfootnote{*Correspondence: \texttt{\href{mailto:vivek.chavan@ipk.fraunhofer.de}{vivek.chavan@ipk.fraunhofer.de}}}

The field of Egocentric Computer Vision has seen increased attention in recent years \cite{plizzari2024outlook, 2012_ego}. This has been catalysed due to the increased mainstream focus on wearable Augmented Reality (AR) and Virtual Reality (VR) devices \cite{Facebook2014Oculus, Apple_vision}. The Computer Vision community has introduced several novel datasets in recent years, with an aim to unlock and explore new challenges and innovations in this area \cite{epic_kitchens, ego4d, egoexo4d}. These large and diverse datasets capture humans in varying everyday scenarios. 

In contrast, we focus on industrial production scenarios. Industry 4.0, or smart manufacturing, focuses on digital transformation of product development, including manufacturing, use, maintenance, and recycling \cite{BMWK2024Industrie40}. There is a significant gap between the current state-of-the-art in Artificial Intelligence (AI) and Computer Vision Research, and its integration into traditional production systems \cite{ind_ml}. The bottleneck often tends to be the digitisation of workflows and the inability to capture the expertise of the Subject-Matter Experts (SMEs) proficiently. 

\begin{figure}[t]
\begin{center}
    \begin{subfigure}{0.48\textwidth}
    \includegraphics[width=\textwidth]{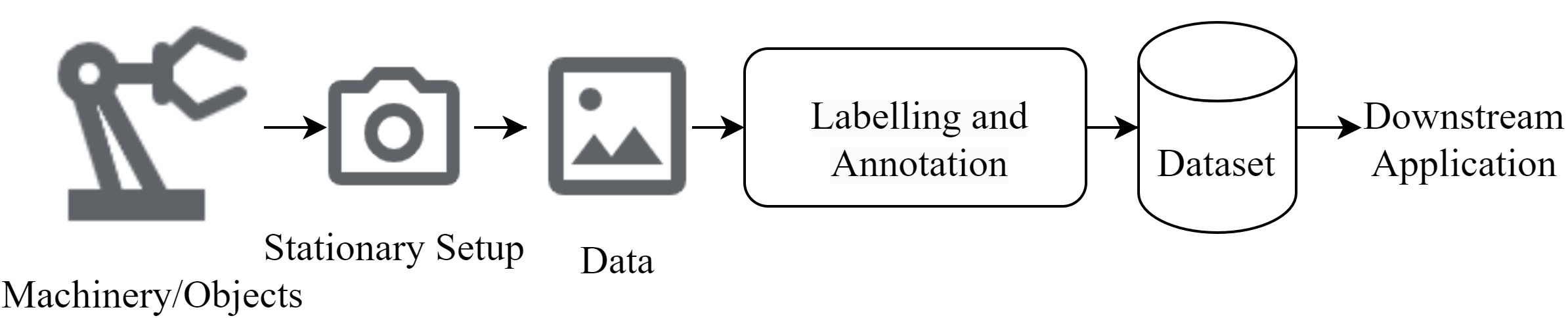}
    \caption{The conventional data collection and labelling approach, involving a fixed, stationary setup.} \label{1a}
      \end{subfigure}
\end{center}
\begin{center}
    \begin{subfigure}{0.48\textwidth}
    \includegraphics[width =\textwidth]{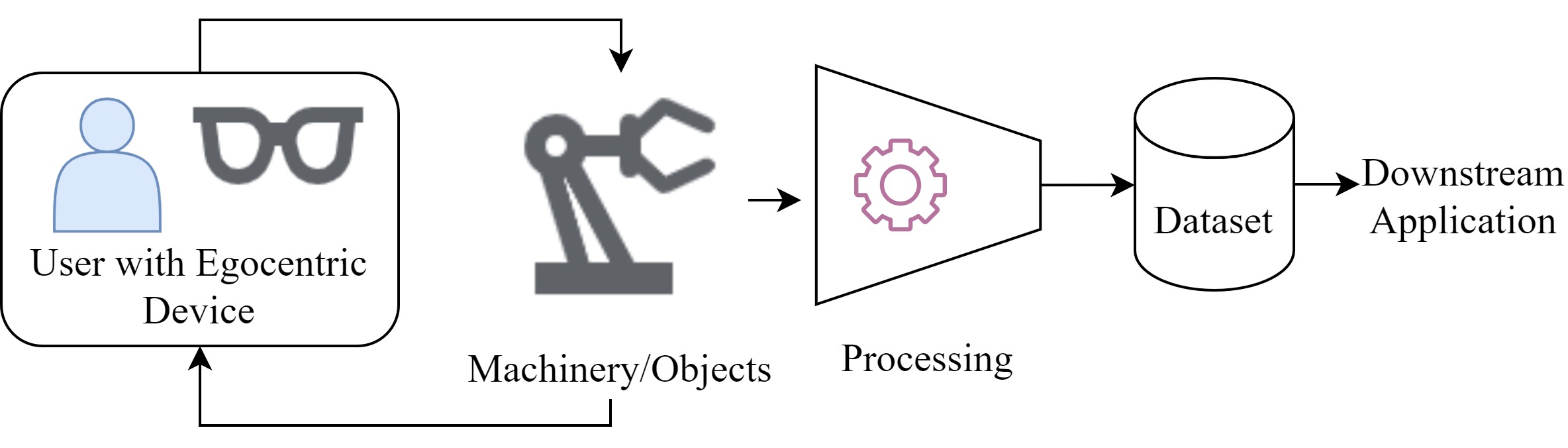}
    \caption{A proposed approach for automated data collection and annotation, where a user describes their observation while interacting with the object. The data is then processed to obtain the labelled dataset.} \label{1b}
      \end{subfigure}
\end{center}
\caption{A comparison of the two approaches. Our work explores the latter.} \label{world_model}
\end{figure}

\begin{figure*}[h]
    \includegraphics[width=\textwidth]{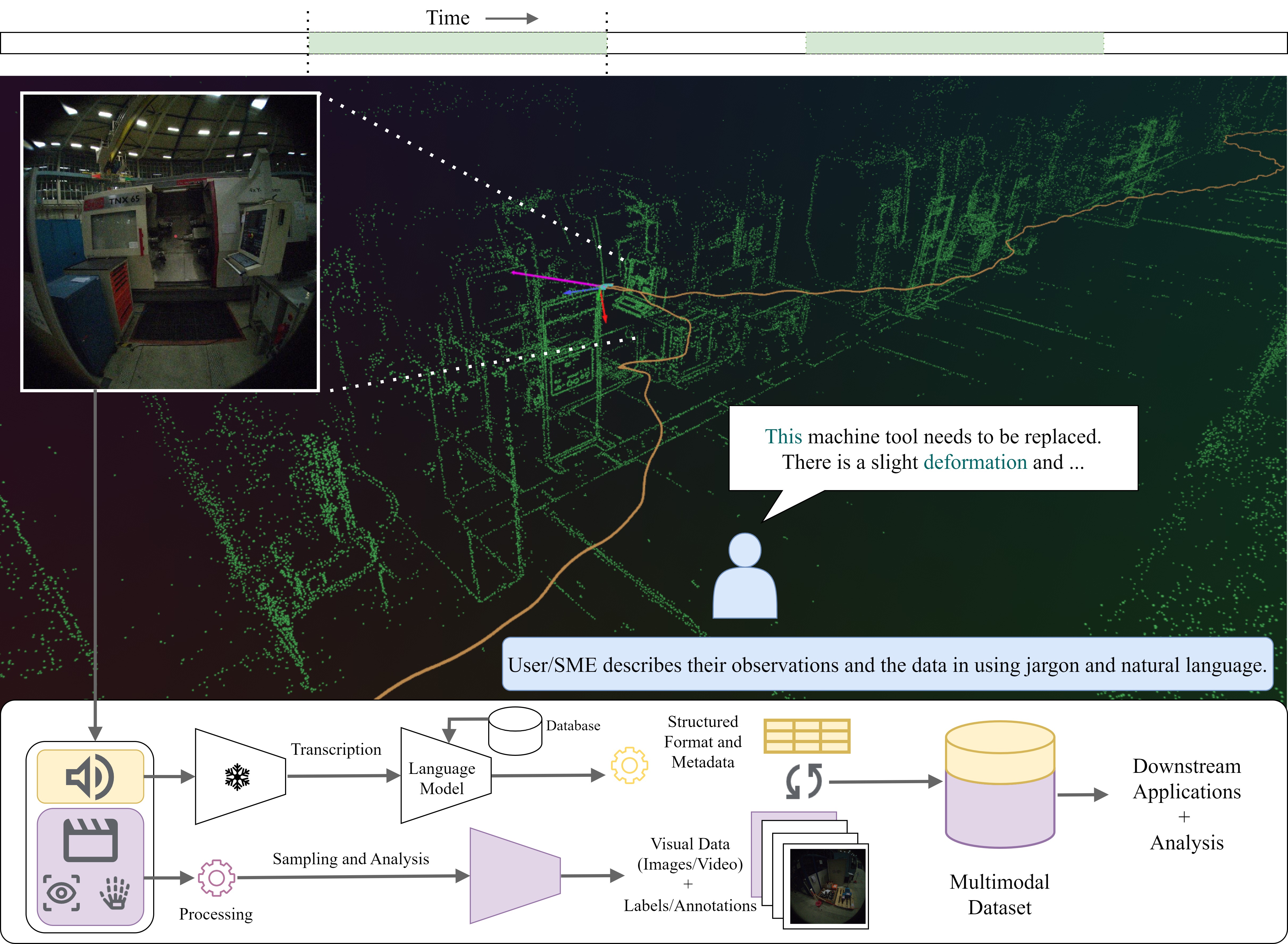}
    \caption{A summary of the proposed pipeline. The User/SME wearing the egocentric device interacts with the object/machinery and documents their observation in natural language. The multimodal dataset is then processed to obtain image/video data, and the transcription, eye-gaze, hand interaction provides the labels and annotations, along with metadata. \textbf{Top: } Point cloud reconstruction example from a use case. \textbf{Bottom: }A conceptualisation of the data processing.}
    \label{aria_concept}
\end{figure*}

The conventional exocentric/allocentric data collection in the industry is summarised in Figure \ref{1a}, which requires careful labelling, annotation and documentation for training AI models or knowledge transfer. In this ongoing research work, we study the use of lightweight egocentric devices for capturing multimodal egocentric data, which is processed via agentic workflow for adding task relevant labels and contextual information to the tasks. This is shown in Figure \ref{1b} and Figure \ref{aria_concept}.


\section{Related Work}
\label{sec:related_work}


\textbf{Egocentric Computer Vision. }Understanding the world from the first-person perspective is intuitive for humans, but poses several challenges for conventional Machine Vision and AI methods \cite{epic_kitchens, ego4d}. Several iterations and configurations of wearable devices have been proposed \cite{wearcomp, ego_stanford, aria_glasses, ego4d}. Such devices enable additional user specific data to be captured alongside visual data, such as eye-gaze, hand pose, voice interaction. Several novel datasets and benchmarks have been released in recent years, which introduce new challenges and research directions for the community. 

\textbf{Industrial Machine Vision. } Traditional image processing has led to several important advancements in the industry, which is further accelerated by deep learning based approaches \cite{ind_ml}. The most important areas of application tend to be classification, object detection, segmentation, anomaly/defect detection.

\section{Methods}
\label{methods}

\newcommand{\mychar}{%
  \begingroup\normalfont
  \includegraphics[height=2\fontcharht\font`\B]{img/ws.png}%
  \endgroup
}

\textbf{Proposed Pipeline. }Figure \ref{aria_concept} shows the planned implementation in an industrial setting. We use the Meta Aria glasses \cite{aria_glasses} as the data capturing device. The multimodal data captured by the user is then processed to extract the most meaningful information about the process, or the machinery. The user guidance via voice serves as the lead indicator for understanding which portion of the continuous stream of the data should be processed. The audio data is processed via a custom language model setup, to obtain structured metadata and labels about the given portion of the stream. The camera stream data, augmented by user interaction (eye-gaze or hands), is processed and synchronised with the audio description data to add annotation and context (e.g. object labels, defects, miscellaneous observations). Additional processed data, such as user trajectory, location and other modalities would also be valuable for adding more context to the captured data.

Current egocentric datasets primarily encompass everyday and outdoor activities, with sparse representation of industry-specific scenarios, necessitating domain-specific data collection. Fine-grained activities like screwing and unscrewing bolts require high-resolution classification due to their visually similar but functionally distinct nature. We plan to open-source our data to encourage the broader research community to address such problems.

\textbf{Challenges. }Industrial Machine Vision often requires controlled settings and high precision image processing. Egocentric data capture cannot fully replace standard digitisation stations and setups. In such cases, egocentric data would augment and assist the user in understanding the workflows and operations. Capturing user guidance via voice may be challenging due to noise, presence of other loud voices or perceived discomfort. In such cases, controlling parts of the user input via hand gestures or other means may be valuable. Additionally, capturing user eye gaze, hand gestures and other personal data poses inherent challenges in such cases. Egocentric AI systems produce large volumes of data, challenging the processing capabilities of on-device hardware. This necessitates adaptive, privacy-centric continual learning strategies and optimisation of data transfer to mitigate compute and bandwidth bottlenecks.

\textbf{Downstream Applications. }We explore diverse applications of egocentric computer vision in industrial settings, aimed at enhancing operational efficiency and accuracy \cite{plizzari2024outlook}. Key applications include improving data collection and annotation through automated processes, enabling precise part recognition and analysis, and facilitating defect and anomaly labelling. Additionally, we investigate scene understanding and action recognition within these environments, offering substantial support for operators through real-time assistance and guidance. The system would also play a crucial role in training and knowledge transfer, ensuring that new and existing employees quickly adapt to evolving industrial demands. Each of these applications underscores the potential of egocentric computer vision to augment traditional industrial operations, making processes more intuitive and interactive.

\textbf{Continual Learning. }NN models often suffer due to catastrophic forgetting when they are retrained on newer data \cite{cat_for, goodfellow_cat_for}. It is necessary to develop systems that learn continuously and perform increasingly better on the most important tasks, while still retaining the knowledge from previous broad scale training. We believe such adaptive and intelligent systems would provide users with a more meaningful and helpful experience over time.  For our work, we focus on a limited number of use cases and working environments. The participants perform similar tasks on different industrial objects multiple times over several weeks. These include tasks involving similarities w.r.t. the objects, working environments, and the actions being performed. 

\begin{figure}[t]
\includegraphics[width=0.48\textwidth]{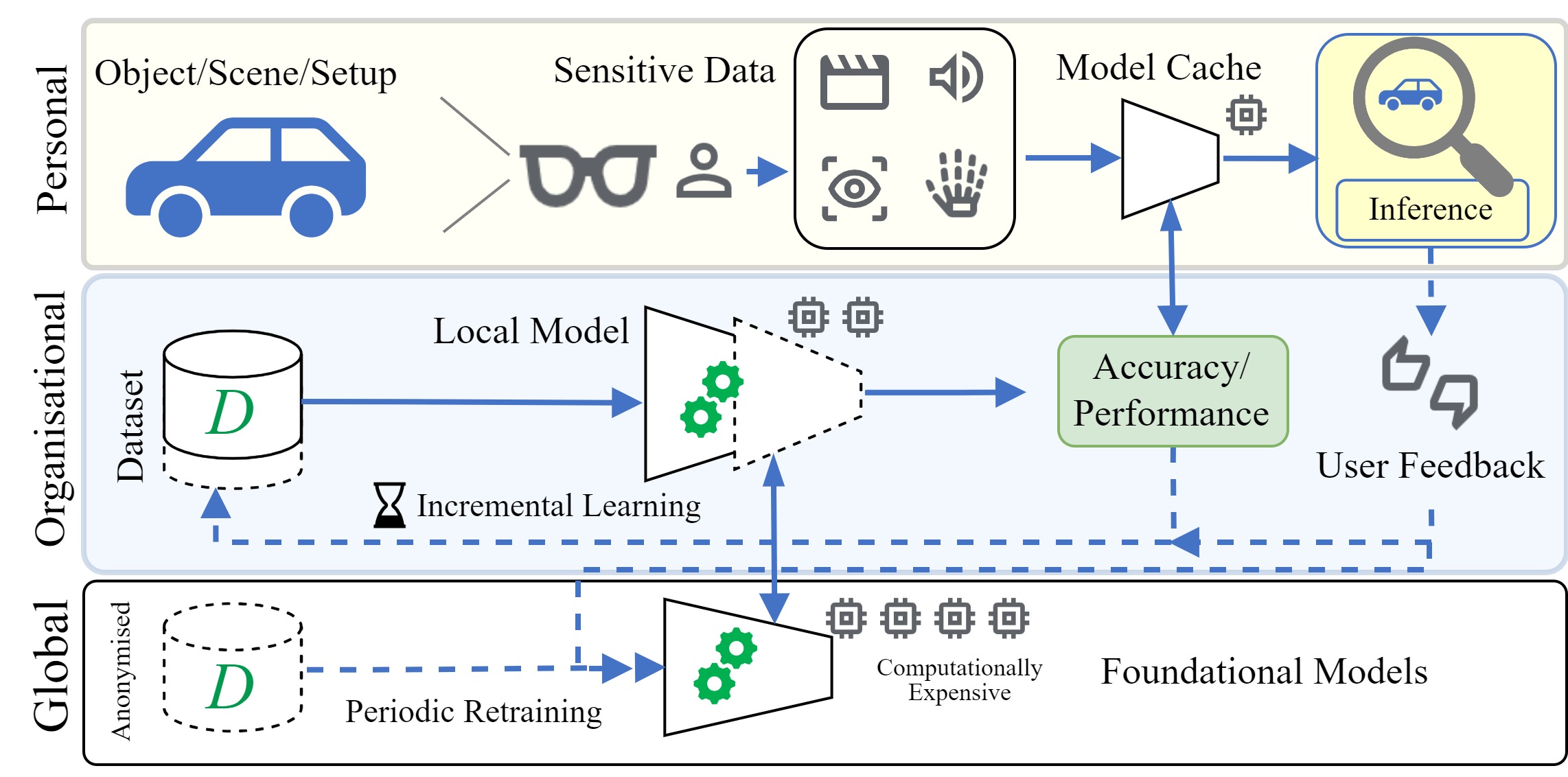}
\caption{A summary of a distributed Continual Learning framework for egocentric applications. The three layers of application include personal (top), organisational (middle) and global (bottom). The most sensitive information is stored and processed by the personal computing setup with limited compute. The organisational layer trains the local models incrementally, which receive user feedback and related data from the egocentric device. The global foundational models require large amounts of data, which could be periodically shared by the organisation (after anonymization and review).}
\label{cl1}
\end{figure}

One of the key challenges with training and continually improving ML models based on egocentric inputs is handling the large amount of multimodal data available to the sensors every second. Moreover, personal data (such as the user eye-gaze and hand pose) or sensitive data (such as confidential office work, conversations at home or work) may not be suitable to be saved and sent for training. Hence, a federated or distributed learning paradigm would be required for handling the data and continually training the models \cite{li2020review, rieke2020future}, as shown in Figure \ref{cl1}.

\section{Summary}

In this extended abstract, we propose an approach for automated data collection and labelling for industrial use cases. The methods and challenges were briefly discussed. This undertaking brings several eccentric benchmarks and tasks, including scene understanding, object detection and tracking, diarisation, action recognition, hand, and eye tracking, among others. We believe such workflows could significantly reduce the efforts required for digitisation and automation, and would improve knowledge transfer between SMEs and trainees, and aid the development of context aware models.

\subsection*{Acknowledgments}
\noindent We thank the Meta AI team and Reality Labs for the Project Aria initiative, including the research kit, the open source tools and related services. The data collection for this study is carried out at the IWF research labs of TU Berlin.
{
    \small
    \bibliographystyle{ieeenat_fullname}
    \bibliography{main}
}


\end{document}